\pdfoutput=1

\documentclass[11pt]{article}

\usepackage[]{acl}

\usepackage{times}
\usepackage{latexsym}

\usepackage[T1]{fontenc}

\usepackage[utf8]{inputenc}

\usepackage{microtype}

\usepackage{booktabs}
\usepackage{multirow}
\usepackage{amsmath}
\usepackage{graphicx,paralist}

\title{PLM-ICD: Automatic ICD Coding with Pretrained Language Models}

\author{Chao-Wei Huang$^{\star\dag}$ \quad
  Shang-Chi Tsai$^{\star}$ \quad
  Yun-Nung Chen$^{\star}$ \\
  $^{\star}$National Taiwan University, Taipei, Taiwan \\
  $^{\dag}$Taiwan AI Labs, Taipei, Taiwan \\
  \texttt{f07922069@csie.ntu.edu.tw\quad y.v.chen@ieee.org}}

\begin{document}
\maketitle
\begin{abstract}
Automatically classifying electronic health records (EHRs) into diagnostic codes has been challenging to the NLP community.
State-of-the-art methods treated this problem as a multi-label classification problem and proposed various architectures to model this problem.
However, these systems did not leverage the superb performance of pretrained language models, which achieved superb performance on natural language understanding tasks.
Prior work has shown that pretrained language models underperformed on this task with the regular fine-tuning scheme.
Therefore, this paper aims at analyzing the causes of the underperformance and developing a framework for automatic ICD coding with pretrained language models.
We spotted three main issues through the experiments: 1) large label space, 2) long input sequences, and 3) domain mismatch between pretraining and fine-tuning.
We propose \textbf{PLM-ICD}, a framework that tackles the challenges with various strategies.
The experimental results show that our proposed framework can overcome the challenges and achieves state-of-the-art performance in terms of multiple metrics on the benchmark MIMIC data.\footnote{The source code is available at \url{https://github.com/MiuLab/PLM-ICD}.}
\end{abstract}

\section{Introduction}
The clinical notes in electronic health records (EHRs) are written as free-form text by clinicians during patient visits.
The notes can be associated with diagnostic codes from the International Classification of Diseases (ICD),  which represent diagnostic and procedural information of the visit.
The ICD codes are a standardized way to encode information systematically and internationally, which could be used for tracking healthcare statistics, quality outcomes, and billing.

While ICD codes provide several useful applications, manually labelling ICD codes has been shown to be very labor-intensive and domain expertise is required~\cite{o2005measuring}.
Hence, automatically assigning ICD codes to clinical notes has been of broad interest in the medical natural language processing (NLP) community.
Prior work has identified several challenges of this task, including the large number of labels to be classified, the long input sequence, and the imbalanced label distribution, i.e., the long-tail problem~\cite{xie2019ehr}.
These challenges make the task extremely difficult, demonstrating that advanced modeling techniques are required.
With the introduction of deep learning models, we have seen tremendous performance improvement on the task of automatic ICD coding~\cite{shi2017towards,xie-xing-2018-neural,mullenbach-etal-2018-explainable,li2020multirescnn,ijcai2020-461,cao-etal-2020-hypercore,liu2021effective,kim2021read,zhou2021automatic}.
These methods utilized convolutional neural networks (CNNs)~\cite{mullenbach-etal-2018-explainable,li2020multirescnn,liu2021effective} or recurrent neural networks (RNNs)~\cite{ijcai2020-461} to transform the long text in clinical notes into hidden representations.
State-of-the-art methods employed a label attention mechanism, i.e., performing attention to hidden representations independently for each label, to combat the extremely large label set~\cite{mullenbach-etal-2018-explainable,ijcai2020-461}.

Recently, pretrained language models (PLMs) with the Transformer~\cite{vaswani2017attention} architecture have become the dominant forces for NLP research, achieving superior performance on numerous natural language understanding tasks~\cite{devlin-etal-2019-bert,liu2019roberta}.
These models are pretrained on large amount of text with various language modeling objectives, and then fine-tuned on the desired downstream tasks to perform different functionalities such as classification~\cite{devlin-etal-2019-bert} or text generation~\cite{radford2019language,2020t5}.

While PLMs demonstrate impressive capabilities across classification tasks, applying PLMs to automatic ICD coding is still not well-studied.
Previous work has shown that applying PLMs to this task is not straightforward~\cite{zhang-etal-2020-bert,pascual2021towards}, and the main challenges being: 
\begin{itemize}
    \item The length of clinical notes exceeds the maximum length of PLMs.
    \item The regular fine-tuning scheme where we add a linear layer on top of the PLMs does not perform well for multi-label classification problems with a large label set.
    \item PLMs are usually pretrained on general-domain corpora, while clinical notes are very medical-specific and the language usage is different.
\end{itemize}
As a result, the performance of PLMs reported in the prior work is inferior to the state-of-the-art models that did not use pre-trained models by a large margin~\cite{pascual2021towards}.
Their best model achieved 88.65\% in terms of micro-AUC, compared with the state-of-the-art 94.9\% from the ISD model~\cite{zhou2021automatic}. This result highlighted that the performance of PLMs on this task was still far from the conventional models.

In this paper, we aim at identifying the challenges met during applying PLMs to automatic ICD coding and developing a framework that could overcome these challenges.
We first conduct preliminary experiments to verify and investigate the challenges mentioned above, and then we propose proper mechanisms to tackle each challenge.
The proposed mechanisms are: 1) domain-specific pretraining for the domain mismatch problem, 2) segment pooling for the long input sequence problem, and 3) label attention for the large label set problem.
By integrating these techniques together, we propose \textbf{PLM-ICD}, a framework specifically designed for automatic ICD coding with PLMs.
The effectiveness of PLM-ICD is verified through experiments on the benchmark MIMIC-3 and MIMIC-2 datasets~\cite{saeed2011multiparameter,johnson2016mimic}.
To the best of our knowledge, PLM-ICD is the first Transformer-based pretrained language model that achieves competitive performance on the MIMIC datasets.
We further analyze several factors that affect the performance of PLMs, including pretraining method, pretraining corpora, vocabulary construction, and optimization schedules.

The contributions of this paper are 3-fold:
\begin{itemize}
    \item We perform experiments to verify and analyze the challenges of utilizing PLMs on the task of automatic ICD coding.
    \item We develop \textbf{PLM-ICD}, a framework to fine-tune PLMs for ICD coding, that achieves competitive performance on the benchmark MIMIC-3 dataset.
    \item We analyze the factors that affect PLMs' performance on this task.
\end{itemize}

\section{Related Work}

\subsection{Automatic ICD Coding}
ICD code prediction is a challenging task in the medical domain. 
Several recent work attempted to approach this task with neural models.
\citet{choi2016doctor} and \citet{baumel2018multi} used recurrent neural networks (RNN) to encode the EHR data for predicting diagnostic results.
\citet{li2020multirescnn} recently utilized a multi-filter convolutional layer and a residual layer to improve the performance of ICD prediction.
On the other hand, several work tried to integrate external medical knowledge into this task.
In order to leverage the information of definition of each ICD code, RNN and CNN were adopted to encode the diagnostic descriptions of ICD codes for better prediction via attention mechanism~\cite{shi2017towards,mullenbach-etal-2018-explainable}.
Moreover, the prior work tried to consider the hierarchical structure of ICD codes~\cite{xie-xing-2018-neural}, which proposed a tree-of-sequences LSTM to simultaneously capture the hierarchical relationship among codes and the semantics of each code.
Also, \citet{tsai-etal-2019-leveraging} introduced various ways of leveraging the hierarchical knowledge of ICD by adding refined loss functions.
Recently, \citet{cao-etal-2020-hypercore} proposed to train ICD code embeddings in hyperbolic space to model the hierarchical structure. Additionally, they used graph neural network to capture the code co-occurrences.
LAAT~\cite{ijcai2020-461} integrated a bidirectional LSTM with an improved label-aware attention mechanism.
EffectiveCAN~\cite{liu2021effective} integrated a squeeze-and-excitation network and residual connections along with extracting representations from all encoder layers for label attention.
The authors also introduced focal loss to tackle the long-tail prediction problem.
ISD~\cite{zhou2021automatic} employed extraction of shared representations among high-frequency and low-frequency codes and a self-distillation learning mechanism to alleviate the long-tail code distribution.
\citet{kim2021read} proposed a framework called Read, Attend, and Code (RAC) to effectively predict ICD codes, which is the current state-of-the-art model on this task.
Most recent models focused on developing an effective interaction between note representations and code representations~\cite{cao-etal-2020-hypercore,zhou2021automatic,kim2021read}.
Our work, instead, is focusing on the choice of the note encoder, where we apply PLMs for their superior encoding capabilities.

\subsection{Pretrained Language Models}
Using pretrained language models to extract contextualuzed representations has led to consistent improvements across most NLP tasks.
Notably, ELMo~\cite{peters-etal-2018-deep} and BERT~\cite{devlin-etal-2019-bert} showed that pretraining is effective for both LSTM and transformer~\cite{vaswani2017attention} models.
Variants have been proposed such as XLNet~\cite{yang2019xlnet}, RoBERTa~\cite{liu2019roberta}.
These models are pretrained on large amount of general domain text to grasp the capability to model textual data, and fine-tuned on common classification tasks.

To tackle domain-specific problems, prior work adapted such models to scientific and biomedical domains, including BioBERT~\cite{10.1093/bioinformatics/btz682}, ClinicalBERT~\cite{alsentzer-etal-2019-publicly}, PubMedBERT~\cite{gu2020domain} and RoBERTa-PM~\cite{lewis-etal-2020-pretrained}.
These models are pretrained on domain-specific text carefully crawled and processed for improving the downstream performance.
The biomedical-specific PLMs reported improved performance on a variety of biomedical tasks, including text mining, named entity recognition, relation extraction, and question answering~\cite{10.1093/bioinformatics/btz682}.

While PLMs achieved state-of-the-art performance on various tasks, applying PLMs to large-scale multi-label classification is still a challenging research direction.
\citet{chang2019taming} proposed X-BERT, a framework that is scalable to an extremely large label set of a million labels.
\citet{lehevcka2020adjusting} showed that the modeling capacity of BERT's pooling layers might be limited for automatic ICD coding.
\citet{pascual2021towards} also demonstrated inferior performance when applying BERT to this task and pointed out several challenges to be addressed.
Specifically, the authors proposed 5 truncation and splitting strategies to tackle the long input sequence problem.
Their proposed \textit{All} splitting strategies is similar to our segment pooling mechanism. However, without the label attention mechanism, the model failed to learn.

\citet{zhang-etal-2020-bert} proposed BERT-XML, an extension of BERT for ICD coding.
The model was pretrained on a large cohort of EHR clinical notes with an EHR-specific vocabulary.
BERT-XML handles long input text by splitting it into chunks and performs prediction for each chunk independently with a label attention mechanism from AttentionXML~\cite{you2019attentionxml}.
The predictions are finally combined with max-pooling.
Our proposed framework, PLM-ICD, shares a similar idea with BERT-XML that we also split clinical notes into segments to compute segment representations.
The main difference is that we leverage an improved label attention mechanism and we use document-level label-specific representations rather than chunk level representations as in BERT-XML.
In Section~\ref{sec:exp}, we demonstrate that PLM-ICD can achieve superior results on the commonly used MIMIC-3 dataset compared with BERT-XML.

\section{Challenges for PLMs}
In this section, we discuss 3 main challenges for PLMs to work on automatic ICD coding and conduct experiments to verify the severity of the challenges.

\begin{table}[t!]
\centering
\begin{tabular}{lccc}
\toprule
\bf Model & \bf Length & \bf Macro-F & \bf Micro-F \\ \midrule
\multirow{2}{*}{LAAT} & 4000 & 9.9      & 57.5     \\
 & 512$^{*}$  & 6.8      & 47.3     \\ \midrule
BERT & 512$^{*}$  &  2.8       & 38.9         \\
\bottomrule
\end{tabular}
\caption{Results of LAAT and BERT on MIMIC-3 with different maximum input lengths (\%). $^{*}$The length is number of words for LAAT and number of tokens for BERT, so their performance cannot directly comparable.}
\label{tab:length}
\vspace{-1mm}
\end{table}

\subsection{Long Input Text}
\label{sec:long_input_text}
Pretrained language models usually set a maximum sequence length as the size of their positional encodings.
A typical value is set to 512 tokens after subword tokenization~\cite{devlin-etal-2019-bert}.
However, clinical notes are long documents which often exceed the maximum length of PLMs.
For instance, the average number of words in the MIMIC-3 dataset is 1,500 words, or  2000 tokens after subword tokenization.

To demonstrate that this is a detrimental problem for PLMs, we conduct experiments on MIMIC-3 where the input text is truncated to 512 words for the strong model LAAT~\cite{ijcai2020-461}, and 512 tokens for BERT.
The results are shown in Table~\ref{tab:length}.
Both models perform worse when the input text is truncated, showing that simple truncation does not work for the long input text problem.
Note that the same trend can be found for other models for ICD coding.
The results reported by \citet{pascual2021towards} also show similar problem where the truncation methods such as \textit{Front-512} and \textit{Back-512} performed much worse than models with longer input context.

\begin{table}[t!]
\centering
\begin{tabular}{lcrc}
\toprule
\bf Model & \bf Codes & \bf Macro-F & \bf Micro-F \\ \midrule
\multirow{2}{*}{LAAT} & 50 & 66.6      & 71.5     \\
 & Full  & 9.9      & 57.5     \\ \midrule
\multirow{2}{*}{BERT} &  50  &  61.5       & 65.4         \\
 & Full & 3.2  & 40.9 \\
\bottomrule
\end{tabular}
\caption{Results of LAAT and BERT on MIMIC-3 with full codes and top-50 codes (\%).}
\label{tab:num_labels}
\vspace{-1mm}
\end{table}

\subsection{Large Label Set}
\label{sec:large_label_set}
Automatic ICD coding is a large-scale multi-label text classification (LMTC) problem, i.e., finding the relevant labels of a document from a large set of labels.
There are about 17,000 codes in ICD-9-CM and 140,000 codes in ICD-10-CM/PCS, while there are 8921 codes presented in the MIMIC-3 dataset.
PLMs utilize a special token and extract the hidden representation of this token to perform classification tasks.
For example, BERT uses a \texttt{[CLS]} token and adds a pooling layer to transform its hidden representation into a distribution of labels~\cite{devlin-etal-2019-bert}.
However, while this approach achieves impressive performance on typical multi-class classification tasks, it is not very suitable for LMTC tasks.
\citet{lehevcka2020adjusting} showed that making predictions based on only the representation of \texttt{[CLS]} token results in inferior performance compared with pooling representations of all tokens, and hypothesized that this is due to the lack of modeling capacity of using the \texttt{[CLS]} token alone.

To examine the PLMs' capability of performing LMTC, we conduct experiments on MIMIC-3 in two settings, \texttt{Full} and \texttt{Top-50}.
The \texttt{Full} setting uses the full set of 8,921 labels, while the \texttt{Top-50} uses the top-50 most frequent labels.
We report the numbers for LAAT directly from~\citet{ijcai2020-461}.
For the BERT model, we use the segment pooling mechanism to handle the long input, which is detailed in Section~\ref{sec:segment_pooling}.
We aggregate the hidden representations of the \texttt{[CLS]} token for each segment with mean-pooling as the document representation.
The final prediction is obtained by transforming the document representation with a linear layer.

The results are shown in Table~\ref{tab:num_labels}.
BERT achieves slightly worse performance than LAAT in the \texttt{Top-50} setting.
However, in the \texttt{Full} setting, BERT performs significantly worse compared with LAAT.
The results suggest that using BERT's \texttt{[CLS]} token for LMTC is not ideal, and advanced techniques for LMTC are required for PLMs to work on this task.

\subsection{Domain Mismatch}
Normally, PLMs are pretrained on large amount of general-domain corpora which contains billions of tokens.
The corpora is typically crawled from Wikipedia, novels~\cite{zhu2015aligning}, webpages, and web forums.
Prior work has shown that the domain mismatch between the pretraining corpus and the fine-tuning tasks could degrade the downstream performance~\cite{gururangan-etal-2020-dont}.

Specifically for the biomedical domain, several pretrained models have been proposed which are pretrained on biomedical corpora to mitigate the domain mismatch problem~\cite{10.1093/bioinformatics/btz682,alsentzer-etal-2019-publicly,gu2020domain,lewis-etal-2020-pretrained}.
These models demonstrate improved performance over BERT on various medical and clinical tasks, showing that domain-specific pretraining is essential to achieve good performance.

\begin{figure*}[t!]
\centering
\includegraphics[width=\linewidth]{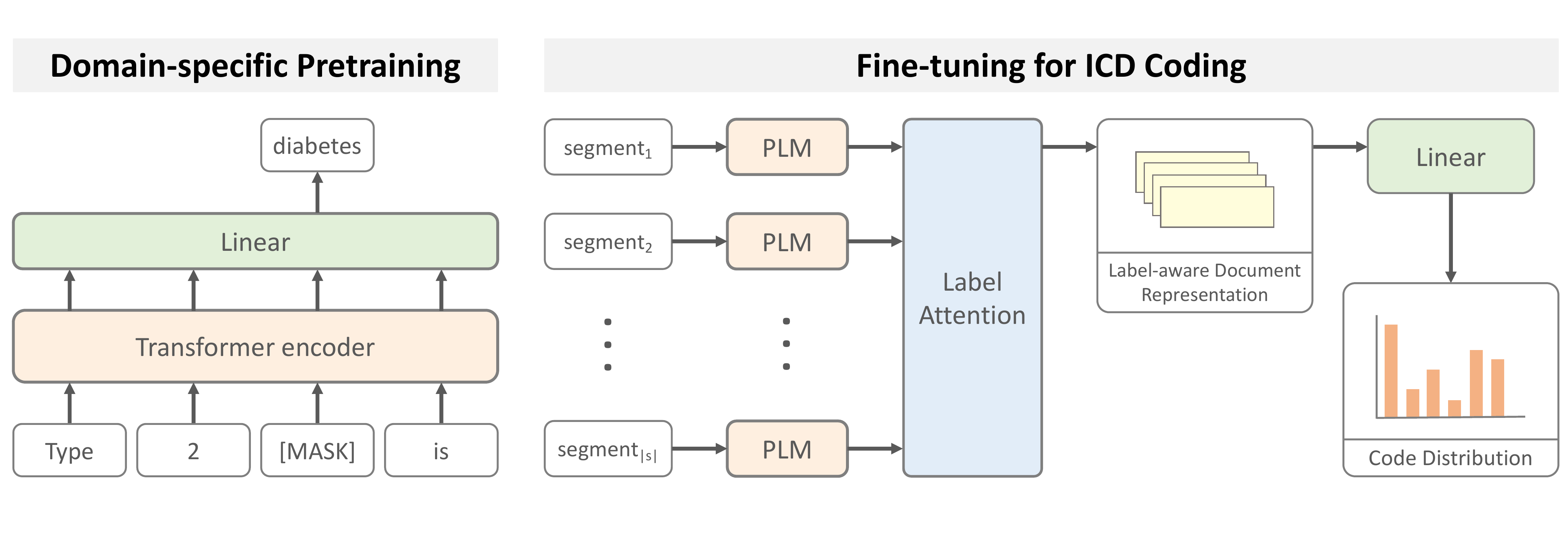}
\vspace{-6mm}
\caption{Illustration of our proposed framework. Left: domain-specific pretraining, where a PLM is pretrained on text from specific domains with a language modeling objective. Right: PLM encodes segments of a document separately, and a label-aware attention mechanism is to aggregate the segment representations into label-aware document representations. The document representations are linear-transformed to predict ICD codes.}
\label{fig:framework}
\vspace{-1mm}
\end{figure*}

\section{Proposed Framework}
The task of ICD code prediction is formulated as a multi-label classification problem~\cite{kavuluru2015empirical,mullenbach-etal-2018-explainable}.
Given a clinical note of $|d|$ tokens $\mathbf{d} = \{t_1, t_2, \cdots , t_{|d|}\}$ in EHR, the goal is to predict a set of ICD codes $\mathbf{y} \subseteq \mathcal{Y}$, where $\mathcal{Y}$ denotes the set of all possible codes.
Typically, the labels are represented as a binary vector $\mathbf{y} \in \{0,1\}^{|\mathcal{Y}|}$, where each bit $y_i$ indicates whether the corresponding label is presented in the note.

The proposed framework \textbf{PLM-ICD} is illustrated in Figure~\ref{fig:framework}.
The details of the components are described in this section.

\subsection{Domain-Specific Pretraining}
Automatic ICD coding is a domain-specific task where the input text consists of clinical notes written by clinicians.
The clinical notes contain many biomedical terms, and understanding these terms is essential in order to assign ICD codes accurately.
While general PLMs are pretrained on large amount of text, the pretraining corpora usually does not contain biomedical text, not to mention clinical records.

In order to mitigate the domain mismatch problem, we propose to utilize the PLMs that are pretrained on biomedical and clinical text, e.g., BioBERT~\cite{10.1093/bioinformatics/btz682}, PubMedBERT~\cite{gu2020domain}, and RoBERTa-PM~\cite{lewis-etal-2020-pretrained}.
These PLMs are specifically pretrained for biomedical tasks and proven to be effective on various downstream tasks.
We take the domain-specific PLMs and fine-tune them on the task of automatic ICD coding.
We can plug-and-play the domain-specific PLMs since their architectural design and pretraining objective are identical to their general-domain counterparts.
This makes our framework agnostic to the type of PLMs, i.e., we can apply any transformer-based PLMs as the encoder.

\subsection{Segment Pooling}
\label{sec:segment_pooling}
In order to tackle the long input text problem described in Section~\ref{sec:long_input_text}, we propose \textbf{segment pooling} to surpass the maximum length limitation of PLMs.
The segment pooling mechanism first splits the whole document into segments that are shorter than the maximum length, and encodes them into segment representations with PLMs.
After encoding segments, the segment representations are aggregated as the representations for the full document.

More formally, given a document $d = \{t_1, t_2, ..., t_{|d|}\}$ of $|d|$ tokens, we split it into $|s|$ consecutive segments $s_i$ of $c$ tokens:
\begin{equation*}
    s_i = \{t_j \mid c \cdot i \leq j < c \cdot (i+1)\}
\end{equation*}
The segments are fed into PLMs separately to compute hidden representations, then concatenated to obtain the hidden representations of all tokens:
\begin{equation*}
    \mathbf{H} = \text{concat} ( PLM(s_1), \cdots, PLM(s_{|s|}))
\end{equation*}
The token-wise hidden representations $\mathbf{H}$ can then be used to make prediction based on the whole document.

\subsection{Label-Aware Attention}
To combat the problem of a large label set, we propose to augment the PLMs with the label-aware attention mechanism proposed by~\citet{ijcai2020-461} to learn label-specific representations that capture the important text fragments relevant to certain labels.
After the token-wise hidden representations $\mathbf{H}$ are obtained, the goal is to transform $\mathbf{H}$ into label-specific representations with attention mechanism.

The label-aware attention takes $\mathbf{H}$ as input and compute $|\mathcal{Y}|$ label-specific representations.
This mechanism can be formulated into 2 steps.
First, a label-wise attention weight matrix $\mathbf{A}$ is computed as:
\begin{align*}
    \mathbf{Z} & = \text{tanh}(\mathbf{VH}) \\
    \mathbf{A} & = \text{softmax}(\mathbf{WZ})
\end{align*}
where $\mathbf{V}$ and $\mathbf{W}$ are linear transforms.
The $i^{th}$ row of $\mathbf{A}$ represents the weights of the $i^{th}$ label, and the softmax function is performed for each label to form a distribution over all tokens.
Then, the matrix $\mathbf{A}$ is used to perform a weighted-sum of $\mathbf{H}$ to compute the label-specific document representation:
\begin{equation*}
    \mathbf{D} = \mathbf{HA}^{\top}
\end{equation*}
where $\mathbf{D}_i$ represents the document representations for the $i^{th}$ label.

Finally, we use the label-specific document representation $\mathbf{D}$ to make predictions:
\begin{equation*}
    \mathbf{p}_i = \text{sigmoid}(\langle \mathbf{L}_i, \mathbf{D}_i \rangle)
\end{equation*}
where $\mathbf{L}_i$ is a vector for the $i^{th}$ label, $\langle \cdot \rangle$ represents inner product between two vectors, $\mathbf{p}_i$ is the predicted probability of the $i^{th}$ label.
Note that the inner product could also be seen as a linear transform with output size 1.
We can then assign labels to a document based on a predefined threshold $t$.

The training objective is to minimize the binary cross-entropy loss $\mathcal{L}(\mathbf{y}, \mathbf{p})$:
\begin{equation*}
    -\frac{1}{|\mathbf{y}|} \sum_{i=1}^{|\mathbf{y}|}
    \Big( \mathbf{y}_i \log \mathbf{p}_i + (1-\mathbf{y}_i) \log (1-\mathbf{p}_i) \Big).
\end{equation*}

\section{Experiments}
\label{sec:exp}
In order to evaluate the effectiveness of our proposed framework, we conduct experiments and compare the results with the prior work.

\subsection{Setup}
We evaluate PLM-ICD on two benchmark datasets for ICD code prediction.
\begin{itemize}
    \item \textbf{MIMIC-2} \quad To be able to directly compare with the prior work~\cite{mullenbach-etal-2018-explainable,li2020multirescnn,ijcai2020-461}, we evaluate PLM-ICD on the MIMIC-2 dataset~\cite{saeed2011multiparameter}.
    We follow the setting from~\citet{mullenbach-etal-2018-explainable}, where 20,533 summaries are used for training, and 2,282 summaries are used for testing.
    There are 5,031 labels in the dataset.
    \item \textbf{MIMIC-3} \quad The Medical Information Mart for Intensive Care III (MIMIC-3)~\cite{johnson2016mimic} dataset is a benchmark dataset which contains text and structured records from a hospital ICU.
    We use the same setting as~\citet{mullenbach-etal-2018-explainable}, where 47,724 discharge summaries are used for training, with 1,632 summaries and 3,372 summaries for validation and testing, respectively.
    There are 8,922 labels in the dataset.
\end{itemize}

The preprocessing is done by following the steps described in~\citet{mullenbach-etal-2018-explainable} with their provided scripts~\footnote{\url{https://github.com/jamesmullenbach/caml-mimic}}.
Detailed training setting is provided in Appendix~\ref{sec:training_details}.

\subsection{Evaluation}
We evaluate our methods with commonly used metrics to be directly comparable to previous work.
The metrics used are macro F1, micro F1, macro AUC, micro AUC, and precision@K, where $K=\{5,8,15\}$.

\begin{table*}[t!]
\centering
\begin{tabular}{l|cc|cc|ccc}
\toprule
 \multirow{2}{*}{\bf Model} & \multicolumn{2}{c}{\bf AUC} & \multicolumn{2}{|c|}{\bf F1} & \multicolumn{3}{c}{\bf P@k} \\
 &\bf Macro & \bf Micro & \bf Macro & \bf Micro &\bf P@5 & \bf P@8 & \bf P@15 \\
\midrule
CAML~\shortcite{mullenbach-etal-2018-explainable} & 89.5 & 98.6 & 8.8 & 53.9 & - & 70.9 & 56.1 \\
DR-CAML~\shortcite{mullenbach-etal-2018-explainable} & 89.7 & 98.5 & 8.6 & 52.9 & - & 69.0 & 54.8 \\
MultiResCNN~\shortcite{li2020multirescnn} & 91.0 & 98.6 & 8.5 & 55.2 & - & 73.4 & 58.4 \\
LAAT~\shortcite{ijcai2020-461} & 91.9 & 98.8 & 9.9 & 57.5 & 81.3 & 73.8 & 59.1 \\ 
JointLAAT~\shortcite{ijcai2020-461} & 92.1 & 98.8 & \bf 10.7 & 57.5 & 80.6 & 73.5 & 59.0 \\
EffectiveCAN~\shortcite{liu2021effective} & 91.5 & 98.8 & 10.6 & 58.9 & - & 75.8 & 60.6 \\
\midrule
PLM-ICD (Ours) & \bf 92.6 \tiny (.2) & \textbf{98.9} \tiny (.1) & 10.4 \tiny (.1) & \textbf{\textit{59.8$^\dag$}}  \tiny (.3) & \textbf{\textit{84.4$^\dag$}} \tiny (.2) & \textbf{\textit{77.1$^\dag$}} \tiny (.2) & \textbf{\textit{61.3$^\dag$}} \tiny (.1) \\
\midrule
\multicolumn{8}{l}{\textit{Models with Special Code Description Modeling}} \\
HyperCore~\shortcite{cao-etal-2020-hypercore} & 93.0 & 98.9 & 9.0 & 55.1 & - & 72.2 & 57.9 \\
ISD~\shortcite{zhou2021automatic} & 93.8 & 99.0 & 11.9 & 55.9 & - & 74.5 & - \\
RAC~\shortcite{kim2021read} & \it 94.8 & \it 99.2 & \it 12.7 & 58.6 & 82.9 & 75.4 & 60.1 \\
\bottomrule
\end{tabular}
\caption{Results on the MIMIC-3 full test set (\%). The best scores among models without special code description modeling are marked in {\bf bold}. The best scores among all models are {\it italic}. The values in the parentheses are the standard variation of runs. $\dag$ indicates the significant improvement with $p < 0.05$.}
\label{tab:mimic3} 
\end{table*}

\begin{table*}[t!]
\centering
\begin{tabular}{l|cc|cc|ccc}
\toprule
 \multirow{2}{*}{\bf Model} & \multicolumn{2}{c}{\bf AUC} & \multicolumn{2}{|c|}{\bf F1} & \multicolumn{3}{c}{\bf P@k} \\
 &\bf Macro & \bf Micro & \bf Macro & \bf Micro &\bf P@5 & \bf P@8 & \bf P@15 \\
\midrule
CAML~\shortcite{mullenbach-etal-2018-explainable}  & 82.0 & 96.6 & 4.8 & 44.2 & - & 52.3 & -\\
DR-CAML~\shortcite{mullenbach-etal-2018-explainable} & 82.6 & 96.6 & 4.9 & 45.7 & - & 51.5 & -\\
MultiResCNN~\shortcite{li2020multirescnn} & 85.0 & 96.8 & 5.2 & 46.4 & - & 54.4 & - \\
LAAT~\shortcite{ijcai2020-461} & 86.8 & \bf 97.3 & 5.9 & 48.6 & 64.9 & 55.0 & 39.7 \\ 
JointLAAT~\shortcite{ijcai2020-461} & \bf 87.1 & 97.2 & \bf 6.8 & 49.1 & 65.2 & 55.1 & 39.6\\
\midrule
PLM-ICD (Ours) & 86.8 \tiny (.2) & \textbf{97.3} \tiny (.1) & 6.1 \tiny (.1) & \textbf{\textit{50.4$^\dag$}} \tiny (.2) & \textbf{\textit{67.3$^\dag$}} \tiny (.2) & \textbf{56.1$^\dag$} \tiny (.2) & \textbf{\textit{39.9}} \tiny (.2) \\
\midrule
\multicolumn{8}{l}{\textit{Models with Special Code Description Modeling}} \\
HyperCore~\shortcite{cao-etal-2020-hypercore} & 88.5 & 97.1 & 7.0 & 47.7 & - & 53.7 & -  \\
ISD~\shortcite{zhou2021automatic} & \it 90.1 & \it 97.7 & \it 10.1 & 49.8 & - & \it 56.4 & - \\
\bottomrule
\end{tabular}
\caption{Results on the MIMIC-2 test set (\%). EffectiveCAN~\shortcite{liu2021effective} and RAC~\shortcite{kim2021read} did not report results on MIMIC-2. The best scores among models without special code description modeling are marked in bold. The best scores among all models are italicized. The values in the parentheses are the standard variation of the runs. $\dag$ indicates that the improvement is statistically significant with $p < 0.05$.}
\label{tab:mimic2} 
\vspace{-1mm}
\end{table*}

\subsection{Results}
We present the evaluation results in this section.
All the reported scores are averaged over 3 runs with different random seeds.
The results of the compared methods are taken directly from their original paper.
We mainly compare our model, PLM-ICD, with the models without special code description modeling.
The performance of models with special code description modeling, i.e., HyperCore, ISD, and RAC, are also reported for reference.

\subsubsection{MIMIC-3}
The results on MIMIC-3 full test set are shown in Table~\ref{tab:mimic3}.
PLM-ICD achieves state-of-the-art performance among all models in terms of micro F1 and all precision@k measures, even though we do not leverage any code description modeling.
All the improvements are statistically significant.
RAC performs best on AUC scores and macro F1.
We note that the techniques proposed by RAC are complementary to our framework, and it is possible to add the techniques to further improve our results.
However, this is out of the scope of this paper.

\subsubsection{MIMIC-2}
The results on MIMIC-2 test set are shown in Table~\ref{tab:mimic2}.
PLM-ICD achieves state-of-the-art performance among all models in terms of micro F1 and all precision@k measures, similar to the results on MIMIC-3.
All the improvements are statistically significant except for P@15.

In sum, these results show that PLM-ICD is generalizable to multiple datasets, achieving state-of-the-art performance on multiple metrics on both MIMIC-3 and MIMIC-2.

\section{Analysis}
This section provides analysis on factors that affect PLM's performance on automatic ICD coding.

\begin{table}[t!]
\centering
\resizebox{\columnwidth}{!}{
\begin{tabular}{clrr}
\toprule
\multicolumn{2}{c}{\bf Model} & \bf Macro-F & \bf Micro-F \\ \midrule
& PLM-ICD &   10.4    & 59.8  \\
(a) & - domain pretraining &  8.9      & 54.2         \\
(b) & - segment pooling &  7.2      & 54.6        \\
(c) & - label attention &  4.6      & 48.0        \\
\bottomrule
\end{tabular}
}
\caption{Ablation results on the MIMIC-3 full test set (\%).}
\label{tab:ablation}
\vspace{-2mm}
\end{table}

\subsection{Ablation Study}
To verify the effectiveness of the proposed techniques, we conduct an ablation study on MIMIC-3 full test set.
The results are presented in Table~\ref{tab:ablation}.

The first ablation we perform is discarding domain-specific pretraining.
In this setting, we use the pretrained \texttt{RoBERTa-base} model as the PLM, and fine-tune it for ICD coding.
As shown in row (a), the performance slightly degrades after discarding domain-specific pretraining.
This result demonstrates that domain-specific pretraining contributes to the performance improvement.

The second ablation we perform is discarding segment pooling.
In this setting, we replace our segment pooling with the one proposed by~\citet{zhang-etal-2020-bert}
They applied label attention and made code predictions for each segment separately, and aggregated the predictions with max-pooling.
As shown in row (b), replacing our segment pooling results in worse performance.
This result indicates that our proposed segment pooling is more effective for aggregating segment representations.

The third ablation is removing the label attention mechanism.
We fall back to the normal PLM paradigm, i.e., extracting representations of the \texttt{[CLS]} token for classification.
This setting is identical to the one described in Section~\ref{sec:large_label_set}, where we aggregate the representation of the \texttt{[CLS]} token for each segment with mean-pooling, and obtain the final prediction by transforming the aggregated representation with a linear layer.
As shown in row (c), removing label attention mechanism results in huge performance degradation.
The micro F1 score degrades by 11.8\% absolute, while the macro F1 score degrades more than half.
This result demonstrates that the label attention mechanism is crucial to ICD coding, which is an observation aligned with the prior work~\cite{mullenbach-etal-2018-explainable}.

\begin{table}[t!]
\centering
\begin{tabular}{lrrr}
\toprule
\bf Model & \bf Macro-F & \bf Micro-F & $\hat{F}$ \\ \midrule
RoBERTa-PM &   10.4    & 59.8 & 1.35  \\
BioBERT &   9.1     &   57.9 & 1.60       \\
ClinicalBERT &   8.8    &  57.8 & 1.60   \\
PubMedBERT & 9.2  &    59.5 & 1.41   \\
\bottomrule
\end{tabular}
\caption{Results with different PLMs on the MIMIC-3 full test set (\%). $\hat{F}$ is the fragmentation ratio.}
\label{tab:plm}
\vspace{-2mm}
\end{table}

\subsection{Effect of Pretrained Models}
While we have shown that domain-specific pretraining is beneficial to ICD coding, we would like to explore which domain-specific PLM performs the best on this task.
We conduct experiments with different PLMs, including BioBERT~\cite{10.1093/bioinformatics/btz682}, ClinicalBERT~\cite{alsentzer-etal-2019-publicly}, PubMedBERT~\cite{gu2020domain}, and RoBERTa-PM~\cite{lewis-etal-2020-pretrained}.

The results are presented in Table~\ref{tab:plm}.
RoBERTa-PM achieves the best performance among the 4 examined PLMs
This result is in line with the reported results on the BLURB leaderboard~\cite{gu2020domain}, which is a collection of biomedical tasks.

We also report the fragmentation ratio, i.e., the number of tokens per word after subword tokenization as~\cite{chalkidis-etal-2020-empirical}.
We observe that the PLMs with vocabulary trained on biomedical texts (RoBERTa-PM and PubMedBERT) perform better than the ones inherited vocabulary from BERT-base (BioBERT and ClinicalBERT).
The framentation ratio also shows that models with custom vocabulary suffer less on the over-fragmentation problem.

\begin{table}[t!]
\centering
\begin{tabular}{lrr}
\toprule
\bf Model & \bf Macro-F & \bf Micro-F \\ \midrule
LAAT &   10.4    & 59.8   \\
CAML &   8.7 &   58.1       \\
BERT-XML &   8.2     &   56.9       \\
\bottomrule
\end{tabular}
\caption{Results with different attention mechanisms on the MIMIC-3 full test set (\%).}
\label{tab:attention}
\vspace{-1mm}
\end{table}

\subsection{Effect of Label Attention Mechanisms}
We conduct experiments with different label attention mechanisms and report the results in Table~\ref{tab:attention}.
We compare the label attention mechanisms from LAAT~\cite{ijcai2020-461}, CAML~\cite{mullenbach-etal-2018-explainable}
 and BERT-XML~\cite{zhang-etal-2020-bert}.
The results show that the label attention used in LAAT is best-suited to our framework.

\begin{table}[t!]
\centering
\begin{tabular}{lrc}
\toprule
\bf Model & \bf Macro-F & \bf Micro-F \\ \midrule
Ours &   10.4    & 59.8   \\
HIER-BERT &   2.8 &   42.7       \\
Longformer &   5.1     &   51.6       \\
\bottomrule
\end{tabular}
\caption{Results with different strategies for tackling the long input problem on the MIMIC-3 full test set (\%).}
\vspace{-1mm}
\label{tab:long_input}
\end{table}

\subsection{Effect of Long Input Strategies}
We also conduct experiments to verify the effect of different strategies for tackling the long input problem.
As shown in Table~\ref{tab:long_input}, our proposed segment pooling outperforms HIER-BERT~\cite{chalkidis-etal-2019-neural} and Longformer~\cite{Beltagy2020Longformer}, demonstrating the effectiveness of our proposed method.

\subsection{Effect of Maximum Length}
We conduct experiments where we alter the maximum length of the documents and segments to explore the different choices of maximum lengths.
The results are shown in Table~\ref{tab:max_length}.

When fixing the maximum length of the documents to 3,072, we observe that longer segments results in better performance until the segment length reaches 128.
Using a longer maximum document length such as 6144 results in slightly better performance.
However, longer sequences require more computation.
Considering the trade-off between computation and accuracy, we set maximum document length to 3,072 and segment length to 128 as our defaults.

\subsection{Effect of Optimization Process}
Similar to the prior work~\cite{sun2019fine}, we also notice that the fine-tuning process is sensitive to the hyperparameters of the optimization process, e.g., batch size, learning rate, and warmup schedule.

With several preliminary experiments conducted on these factors, we observe that the learning rate and the warmup schedule greatly affects the performance.
When we reduce learning rate to 2e-5, the model performs 3\% worse than using the default parameters in terms of micro F1.
The warmup schedule is crucial in our framework.
When we use constant learning rate throughout training, the model performs about 4\% worse.
We do not observe clear difference between different scheduling strategies.

\begin{table}[t!]
\centering
\begin{tabular}{crcc}
\toprule
\bf Max  & \bf Segment  & \multirow{2}{*}{\bf Macro-F} & \multirow{2}{*}{\bf Micro-F} \\ 
\bf Length & \bf Length  & & \\
\midrule
6144 & 128 & 9.2 & 60.0 \\
3072 & 256 & 9.4 & 59.2 \\
3072 & 128 & 9.2 & 59.6 \\
3072 & 64 & 8.2 & 59.3 \\
3072 & 32 & 6.9 & 57.8 \\
\bottomrule
\end{tabular}
\caption{Results with different maximum lengths on the MIMIC-3 full dev set (\%).}
\label{tab:max_length}
\vspace{-2mm}
\end{table}

\subsection{Best Practices}
With the above analyses, we provide a guideline and possible future directions for applying PLMs to ICD coding or tasks with similar properties:
\begin{compactitem}
    \item With the input length exceeding the maximum length of PLMs, segment pooling can be used to extract representations of all tokens. PLMs with longer input length or recurrence could be explored in the future.
    \item The representation of the \texttt{[CLS]} token might be insufficient when dealing with LMTC problems. A label attention mechanism could be beneficial in such scenarios.
    \item The pretraining corpora plays an important role for domain-specific tasks.
    \item The hyperparameters of the optimization process greatly affect the final performance, so trying different parameters is preferred when the performance is not ideal.
\end{compactitem}

\section{Conclusion}
In this paper, we identify the main challenges of applying PLMs on automatic ICD coding, including the long text input, the large label set and the mismatched domain.
We propose \textbf{PLM-ICD}, a framework with PLMs that tackles the challenges with various techniques.
The proposed framework achieves state-of-the-art or competitive performance on the MIMIC-3 and MIMIC-2 datasets.
We then further analyze factors that affect PLMs' performance.
We hope this work could open up the research direction of leveraging the great potential of PLMs on ICD coding.

\section*{Acknowledgements}

We thank reviewers for their insightful comments.
This work was financially supported from the Young Scholar Fellowship Program by Ministry of Science and Technology (MOST) in Taiwan,
under Grants 111-2628-E-002-016 and 111-2634-F-002-014.

\bibliography{anthology,custom}
\bibliographystyle{acl_natbib}

\appendix

\section{Training Details}
\label{sec:training_details}
We take the pretrained weights released by original authors without any modification.
For the best PLM-ICD model, we use RoBERTa-base-PM-M3-Voc released by~\citet{lewis-etal-2020-pretrained}.
During fine-tuning, we train our models for 20 epochs. 
\texttt{AdamW} is chosen as the optimizer with a learning rate of $5e-5$.
We employ a linear warmup schedule with 2000 warmup steps, and after that the learning rate decays linearly to 0 throughout training.
The batch size is set to 8.
All models are trained on a GTX 3070 GPU.
We truncate discharge summaries to 3072 tokens due to memory consideration, and the length of each segment $c$ is set to 128.
The validation set is used to find the best-performing threshold $t$, and we use it to perform evaluation on the test set.

\end{document}